\documentclass[journal]{IEEEtran}

\usepackage{epstopdf}
\ifCLASSINFOpdf
   \usepackage[pdftex]{graphicx}
   \graphicspath{{C:../pdf/}{C:../jpeg/}{C:../png/}}
   \DeclareGraphicsExtensions{.pdf,.jpeg,.png}
\else
  \usepackage[dvips]{graphicx}
  \graphicspath{{C:../eps/}}
  \DeclareGraphicsExtensions{.eps}
\fi
\usepackage{amsmath}
\makeatletter

\newcommand{\Rmnum}[1]{\expandafter\@slowromancap\romannumeral #1@}
\makeatother
\usepackage{bm}
\usepackage{colortbl}
\usepackage{makecell}
\usepackage{multirow}
\usepackage{amssymb}
\usepackage{float}
\usepackage[colorlinks, linkcolor=black, anchorcolor=black, citecolor=black]{hyperref}
\usepackage{threeparttable}
\usepackage{array}
\usepackage{amsmath}
\usepackage{graphicx}

\usepackage{color}
\usepackage{cite}

\usepackage[ruled]{algorithm2e}

\usepackage{algpseudocode}
\usepackage{graphics}
\usepackage{epsfig}

\usepackage{booktabs}
\usepackage{threeparttable}

\begin{document}

\title{Life-long Learning and Testing for Automated Vehicles via Adaptive Scenario Sampling as A Continuous Optimization Process}

\author{Jingwei Ge, Pengbo Wang, Cheng Chang,  Yi Zhang,~\IEEEmembership{Senior Member,~IEEE}, Danya Yao, Li Li,~\IEEEmembership{Fellow,~IEEE}
\thanks{This work was supported in part by the National Key Research and Development Program of China under Grant 2021YFB2501200. (\emph{Corresponding author: Li Li, Yi Zhang})}
\thanks{Jingwei Ge, Pengbo Wang, Cheng Chang, and Danya Yao are with the Department of Automation, Tsinghua University, Beijing 100084, China .}
\thanks {Yi Zhang is with Department of Automation, Beijing National Research Center for Information Science and Technology (BNRist), Tsinghua University, Beijing 100084, China; Tsinghua-Berkeley Shenzhen Institute (TBSI), Shenzhen 518055, China; Jiangsu Province Collaborative Innovation Center of Modern Urban Traffic Technologies, Nanjing 210096, China. }

\thanks{Li Li is with the Department of Automation, BNRist, Tsinghua University,
Beijing 100084, China (e-mail: li-li@tsinghua.edu.cn).}}

\maketitle
\begin{abstract}

Sampling critical testing scenarios is an essential step in intelligence testing for Automated Vehicles (AVs). However, due to the lack of prior knowledge on the distribution of critical scenarios in sampling space, we can hardly efficiently find the critical scenarios or accurately evaluate the intelligence of AVs. To solve this problem, we formulate the testing as a continuous optimization process which iteratively generates potential critical scenarios and meanwhile evaluates these scenarios. A bi-level loop is proposed for such life-long learning and testing. In the outer loop, we iteratively learn space knowledge by evaluating AV in the already sampled scenarios and then sample new scenarios based on the retained knowledge. Outer loop stops when all generated samples cover the whole space. While to maximize the coverage of the space in each outer loop, we set an inner loop which receives newly generated samples in outer loop and outputs the updated positions of these samples. We assume that points in a small sphere-like subspace can be covered (or represented) by the point in the center of this sphere. Therefore, we can apply a multi-rounds heuristic strategy to move and pack these spheres in space to find the best covering solution. The simulation results show that faster and more accurate evaluation of AVs can be achieved with more critical scenarios.

\end{abstract}

\begin{IEEEkeywords}
Intelligence testing, Senario generation, Life-long learning and testing, Optimization, Automated vehicles.
\end{IEEEkeywords}
\IEEEpeerreviewmaketitle
\section{Introduction}

\IEEEPARstart{I}{ntelligence} testing is a mandatory step in determining that Automated Vehicles (AVs) are trained sufficiently well to be mass-produced \cite{10316374}\cite{10122127}\cite{10470374}\cite{chang2023bev}\cite{zhang2022systematic}. The implementation of testing is mainly carried out in testing scenarios, where AV is required to handle various testing tasks and then be evaluated\cite{winkelmann2023vectorized}\cite{li2016intelligence}\cite{10130026}\cite{yi2021technologies}\cite{7974888}\cite{li2019parallel}.
\par Many studies \cite{wang2021harmonious}\cite{9526613}have noted that AVs rarely make mistakes in the scenarios where they have been trained or experienced. For certain scenarios that it has not experienced, AV is not always able to pass\cite{TENG}. In testing, it is expected to identify these critical scenarios as soon as possible through sampling and generation. So that we can, in turn, upgrade or modify the AV based on the testing results, realizing a closed loop of learning and testing\cite{li2020theoretical}.
\par However, most of the current intelligent algorithms for training AVs are unexplainable and black-box, which makes it difficult to know what specific scenarios are critical for an AV\cite{feng2020testing}\cite{feng2021intelligent}. Obviously, testing AV in all possible scenarios is unrealistic. Instead, we should speculate the potentially critical scenarios based on the AV evaluation results (or scores) in known scenarios.
\par To realize this, the following two questions revealed in \cite{li2018artificial} need to be answered:
\begin{figure*}
	\centering
	\includegraphics[width=5.1in,keepaspectratio]{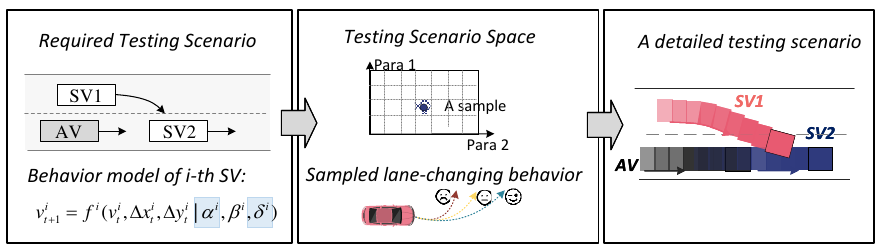}
	\caption{Scenario generation using parameterized space. The static environment of scenario is constructed and the parameters of SVs models, like $\alpha^i$ and $\beta^i$ are extracted; Then the parameters are used to construct scenario sampling space and the point in space represents a scenario; The sampled point finally contributes to a detailed scenario for AV. }
	\label{Fig1}
\end{figure*}
\par \textit{Q}1. \textit{"How to define testing scenarios?"} A testing scenario typically involves a static environment in a segment of space and interactions between dynamic traffic participants, i.e., Surrounding Vehicles (SVs) in this paper, over a period of time. Some literature discretizes the testing scenario and defines it as a spatio-temporal segment consisting of snapshots arranged by sampling time \cite{10401017}\cite{10190119}\cite{chang2022metascenario}\cite{10008082}. Each snapshot describes the positional state of each SVs at a specific sampling moment\cite{corso2019adaptive}\cite{koren2021finding}. However, the testing scenarios are complex, and this approach is prone to what is known as the dimensionality catastrophe\cite{feng2021intelligent}. Therefore, in this paper, we opt for another type of scenario definition, namely parameterized scenarios, as shown in Fig.\ref{Fig1}. Important scenario parameters are extracted as variables and the scenario parameters space (we refer to this as space for simplicity) is constructed accordingly. The coordinates of each point in space represent the parameters that can be used to construct various detailed scenarios. A small number of these parameters can capture the reasonable and rich spatial-temporal relationships that arise from vehicle interactions in the scenarios\cite{10415067}\cite{ding2023survey}.
\par \textit{Q}2. \textit{"How to find critical scenarios as fast as possible?"} Some studies assumed that AV may not work well in the rare scenarios and thus focused on naturalistic driving data. \cite{zhao2016accelerated}\cite{ding2021multimodal}. This involves assuming the frequency of occurrence of different scenarios in the dataset as a form of distribution. Monte Carlo search is then used in combination with importance sampling to search for a fitted distribution \cite{chen2020evaluate}. Through this, scenarios with low frequency of occurrence can be sampled as critical scenarios to accelerate testing efficiency. However, \cite{li2020theoretical} points out that considering both the frequency and criticality of scenarios at the same time often makes it difficult to eliminate estimation and approximation errors. Additionally, optimization-based testing methods are also worth exploring further. For instance, in \cite{li2024few}, the authors introduce the few-shot testing framework. This framework optimizes the Monte Carlo error to identify challenging scenarios with a minimal number of samples. By doing this, datasets with large-scale scenarios is covered equivalently. However, their method still relies on natural distribution of naturalistic driving data. In contrast, \cite{10415067} proposes the dynamic testing for AV. This approach highlights an unknown distribution of critical scenarios, rather than a given distribution. It describes sampling as an optimization problem for space coverage considering a limited number of samples. The discrepancy of the samples in space is used as an optimization objective to cover the space uniformly, thus achieving a fast understanding of the sampling space.

\begin{figure*}[htb]
	\centering
	\includegraphics[width=5.1in,keepaspectratio]{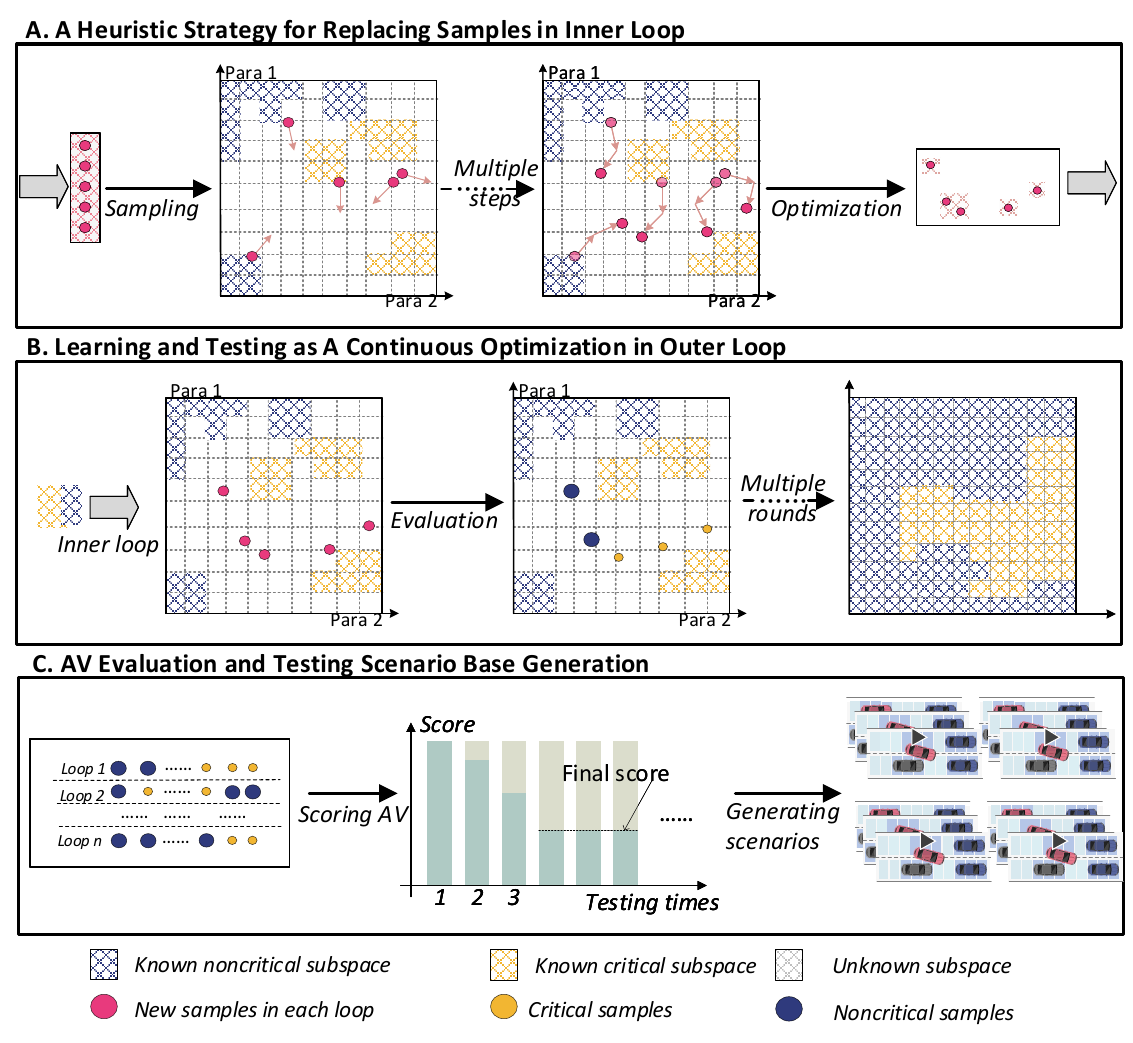}
	\caption{Important processes in life-long learning and testing.}
	\label{Fig2}
\end{figure*}
\par However, it is important to consider that with the scenario scale increases, it is difficult to obtain the real distribution of scenarios in high-dimensional space \cite{wang2019enabling}\cite{xia2023understanding}\cite{zhu2023learning}. Without a clear understanding of the scenario distribution, it is usually very difficult to cover the critical subspaces \cite{li2018artificial}\cite{lamb2016two}. Besides, up to now, no method has been claimed to be well enough to cover the entire high-dimensional space. Thus, method using only few samples or trying to enumerate all samples to cover the space is hard to be realized. In addition, the intelligence (or capacity) of AV may change significantly as learning progresses. When the measured capability changes, the distribution of challenging scenarios will also change and then the testing scenarios base will be updated to deal with the enhancement of AV \cite{li2018artificial}. 
\par Therefore, we believe a completely different scheme than the above approaches and emphasize that life-long testing is necessary, as pointed out in \cite{li2020theoretical}.
To realize this, we further propose a new scheme to conduct multiple and continuous tests on AV. The goal is to continuously evaluate the intelligence of AV by finding more critical scenarios. The criticality of scenario is determined by whether AV smoothly and successfully passes it. When AV encounters a challenging event, its score decreases, also indicating that the scenario is critical to it. 
\par Thus the testing can be regarded as a continuous optimization process to minimize the score of AV. When the samples cover all the space, the testing stops and the AV is evaluated with the final score.  Based on this, a loop, namely outer loop, is set where we continuously generate new samples, evaluate AV in them, and then use the evaluation results as prior knowledge to assist the next round until the termination condition is reached.
\par Different from the previous method, we believe although it needs to be synchronized to quickly find critical scenarios and cover the sampling space, they are difficult to have both.  But in life-long learning and testing, scenarios that AV has not been experienced will always be sampled and generated from the unknown sampling space. 
\par Further, it leads to the third question: \textit{Q}3. \textit{"How to cover the unknown space as much as possible in each round based on prior knowledge?"}. 
\par \textit{Q}2 and \textit{Q}3 are two related but different queations. To answer \textit{Q}3, we assume that the testing results of a point in space, i.e., a sample, can represent the testing results of other points in a sufficiently small subspace around it. Such a subspace can be called the representable subspace of the point. The representable subspaces of all selected points should maximize the coverage of unknown space to avoid waste of testing resources. However, in the case of multiple rounds of sampling, the known space may be complex and irregular (such as the space full with randomly generated points), resulting in the subspace represented by the new points not only easy to overlap with the known space, but also overlap with each other. 
\par Thus, we set another loop, namely inner loop, embed the outer loop mentioned above. We formulate it into a sub-optimization problem which minimizes the overlapping range of these representable subspaces by reasonably reallocating the positions of the points generated in each outer loop. The solution to sub-optimization problem is also helpful to quickly find the optimal answer to the optimization in outer loop.
\par To solve the sub-optimization problem, we propose an adaptive sampling algorithm with a multi-rounds heuristic strategy. It assumes such a representative subspace as a multi-dimensional sphere, the center of the which is the point representing this subspace. Thus, the sphere radius corresponds to the degree of criticality of the sample. The subspace occupied by this sphere is the coverage of the sample point. Further, in every inner loop, we put new sample spheres in space (generated in outer loop) and constantly move them to reduce the overlap of spheres. In order to calculate the moving direction of the sphere, we regard the sample as a sphere with repulsive force, and the sample sphere will be repelled by the surrounding sphere. We design the sphere to move away from other spheres along the direction of the resultant force. The movement continues until the end of the iteration.
\par In addition, another advantage of sphere assumption is that we can make a unified geometric interpretation of the previous papers more vividly. For example, standard Monte Carlo is equivalent to putting spheres of equal size into space randomly, which faces the clumping problem because the sample spheres overlap.
\par For better explanation, the following architecture is as follows: Section II provides preliminaries for this paper. In Section III, the overview on life-long learning and testing will be given, along with the construction of the two proposed optimization problems. Section IV presents the details on the adaptive sampling while simulation results are given in Section V proving the superiority of the proposed scheme.

\section{Preliminaries}
\label{sec2}

\par The basic testing process has three main steps: testing preparation, scenario generation and AV evaluation. Among them, testing preparation includes the necessary testing requirements, testing metrics or functionalities selection, testing tasks design, etc. Next, the concepts and roles of scenarios generation and AV evaluation will be described respectively.

\subsection{Description of Scenario Generation Based on Parameterized Space}
\par Testing scenario is the main carrier of testing implementation, in which the AV processes various tasks and accepts evaluation \cite{althoff2018automatic}\cite{tenbrock2021conscend}\cite{lin2022rule}. Specifically, a scenario refers to a segment in a temporal and spatial context. It reflects the interaction between AV and the surrounding environment (mainly SVs) \cite{winkelmann2023vectorized}\cite{li2021scegene}\cite{koren2021finding}\cite{ge2023autonomous}. Mathematically, testing scenario is defined by a series of vehicle behavior-related variables and the set of scenarios is defined as $\boldsymbol{X}=\left\{\boldsymbol{x}_{i}=(x_{i,1},...,x_{i,D})| x \in R, i=1,2...\right\}$, where $\boldsymbol{x}_{i}$ is the coordinate of any point in space, $D$ is the space dimension, and $\boldsymbol{x}_{i}$ is $i$-$th$ sample. In this paper, we assume that the important parameters of the scenario are all about the behavior of SVs. That is, we only focus on the interactions between AV and SVs rather than the influence from static environment. Based on this assumption, we construct SVs$'$ driving behavior models, the parameters of which are used as the dimensions of the space, like \cite{10415067}.

\par It can be seen that the points in space are mapped to the detailed scenarios, and the spatial position coordinate of a point represent the variable values of the scenario corresponding to one sample. Besides, when two points in space are close enough, we assume that the difference between them can be ignored. That is, the two scenarios formed by them share the same  criticality. We define the set of similar points around the point as the representative subspace of the point. Since the discussion of scenario similarity is not the focus of this article, this part will not be further discussed. Related research can be seen in \cite{chang2022metascenario} for your interests. This representative subspace can also be viewed as the covering area of a point. Supposing there must exist $\left\{\varepsilon_{i,d}|d=1,...,D\right\}$ for all points in $\chi(\boldsymbol{x}_i) =\left\{|x'_d-x_{i,d}|\leq\varepsilon_{i,d}|d=1,2,...,D\right\}$ that share similar testing results, we define this area is the coverage range of point $\boldsymbol{x}_{i}$ and the size of this area is $Volume(\chi(\boldsymbol{x}_i))$, where $i$ is the label for one sample. 

\par The subspace represented by the already sampled points $\boldsymbol{X}$ is defined as a known subspace. The repeated coverage of the known space will undoubtedly increase the test cost. While the remaining unexplored subspace is defined as an unknown subspace, in which the distribution of critical scenarios is still unknown. Therefore, it is inevitable to retain spatial knowledge by sampling points and covering unkonwn subspaces.
\subsection{Definition of Critical Scenario and AV Evaluation}
\par As mentioned above, though it is difficult to cover the entire space, we still need to find enough critical scenarios as soon as possible to evaluate AV. In this paper, the criticality of the scenario is related to the results whether AV could pass it. So a critical scenario is defined as the scenario that AV passes when it fails to complete the testing task.
\par The testing metric, relating to capability of AV, is the standard to assess the completion of the task, which can be $0/1$ binary or numerical value. For example, AV is expected to have driving intelligence that can improve traffic safety and efficiency. There are many metrics that can be used. We choose TTC to be a safety related metric in this paper since it's popular and often-used in many studies. While deceleration value can be used as an efficiency related metric. If the metric is less than the preset threshold, it means that AV fails to complete the task\cite{li2016intelligence}. We use a indicator function to express whether scenario $\boldsymbol{x}$ is challenging:

\begin{equation}
	\begin{gathered}
	 I\left(\boldsymbol{x}\right)=\left\{\begin{array}{c}{1, if \quad TTC \leq TTC_\theta \quad or \quad a \leq -|a_\theta|}\\{0, otherwise}\end{array}\right. 
	\end{gathered}
\end{equation}
where $TTC_\theta$ and $a_\theta$ are the thresholds.

\section{Life-long learning and testing}

\subsection{The Pipeline of the Life-long Learning and Testing}
\par Fig.\ref{Fig2} is the flow chart of life-long learning and testing. Testing requirements are used as input, including concerned testing metrics, designed infinite (or finite) testing tasks, etc. The output includes two parts: the final score of AV and the generated testing scenario library.
\par
\begin{figure}
	\centering
	\includegraphics[width=2.2in,keepaspectratio]{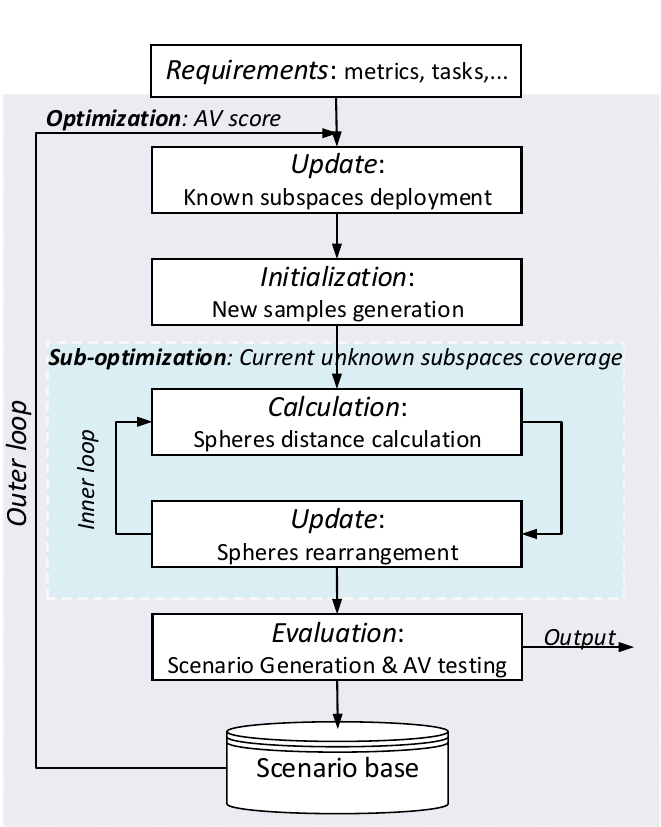}
	\caption{Pipeline of the proposed scheme.}
	\label{Fig3}
\end{figure}

 The main skeleton of the process consists of two loops, outer and inner. The inner loop maximizes the coverage of the currently unknown subspace by sampling a fixed number of points. In the outer loop, the evaluation results and these scenarios will be stored in the scenario base as space knowledge, and these past knowledges will be used as prior knowledge to assist the next inner loop, as shown in Fig.\ref{Fig3}.
\par The importance of life long learning and testing is reflected in three aspects. 1. We can quickly learn the knowledge of the unknown space, so as to speed up finding more critical scenarios. 2. The idea of life long can be well compatible with AV whose ability can change. This type of AV includes not only active learning ones, but also passive upgraded or modified ones. 3. In the spiral of continuous learning and testing, we can correctly assess the current intelligence level of AV.

\subsection{Testing as Optimization in Outer Loop}

\par Though it is difficult to find all critical samples and we do not know what the adequate sample size is, we still have to evaluate the AV as accurately as possible.
\par To realize this, we define the score of an AV to be dependent on the scenarios it experiences, especially the scenarios that can make the AV perform poorly. The more critical scenarios the AV encounters, and the more poorly it performs, the lower the score of the AV is. From this, we define the score

\begin{equation}
	Score(\boldsymbol{X}) = S - \sum I(\boldsymbol{x}), \boldsymbol{x} \in \boldsymbol{X}
\end{equation}
where $S$ is a constant which represents the max of score. The constant can be set to the total number of generated scenarios.
\par So the testing of AV is to minimize the score of AV by generating critical scenarios, which is equal to find the solution for the optimization problem
\begin{equation}
	\min_{\boldsymbol{X}} Score(\boldsymbol{X})	\\
\end{equation}

\par However, since we cannot enumerate all the possible scenarios,  we define a reference indicator to  help us decide when to stop testing. In this paper, we suggest using the proportion of space covered by the obtained samples as such an indicator. 
\begin{equation}
	CRate(\boldsymbol{X}) =\frac{Volume(\chi(\boldsymbol{X}))}{V_{space}}
\end{equation}
where $Volume(\chi(\boldsymbol{X}))$ is the volume of the subspace $\bigcup\chi(\boldsymbol{x}_i)$. 
\par $CRate(\boldsymbol{X})$ is dependent on the size of the subspace represented by the samples and the number of the obtained samples. If the volume of the subspace represented by a sample is small, more samples are required to fill the space. We will elaborate on this in the simulations.
\par From the above, we can infer that the entire mechanism aims to achieve two objectives: reducing the AV score and increasing the $	CRate(\boldsymbol{X})$. It is evident that accomplishing both goals is challenging, as highlighted in $Q$2 and $Q$3, which are related but distinct issues.
\par Therefore, outer loop is proposed with the $Score(\boldsymbol{X})$ as the primary optimization objective and  $CRate(\boldsymbol{X})$ as a constraint. The core of the outer loop is the optimization search through batch sampling. The evaluation of AV is expected to be a gradual process, where scenarios with poor AV performance are continuously identified and the  $Score(\boldsymbol{X})$ is reduced until the coverage stopping condition is met. 

\par The outer loop generates a specific number of samples at each iteration and evaluates the AV based on the samples of the inner loop outputs. The final AV score is the total score obtained after testing the scenarios generated in all loop rounds.
Thus, the score function can be rewritten as,
\begin{equation}
	Score(\boldsymbol{X})=S-\sum_{r}\sum_{i=1}^{N(r)}I(\boldsymbol{x}_i)
\end{equation} 
where $\boldsymbol{x}_i \in \bigcup_{r}\boldsymbol{X}^{(r)}$.
\par Then the optimization for outer loop is

\begin{equation}\begin{gathered}\label{eq4}
	\min_{\boldsymbol{X}} S-\sum_{r}\sum_{i=1}^{N(r)}I(\boldsymbol{x}_i)	\\
	s.t. CRate(\boldsymbol{X})< CRate_{\theta} \\
	s.t. \boldsymbol{X}=\left\{{\boldsymbol{x}_i|i=1,2,...,\sum_{r}N(r)}\right\}\\
	s.t. N(r),r < \infty
	\end{gathered}
\end{equation}
where $CRate_{\theta}$ is the thershold for covering area and $N(r)$ is the total number of samples in $r$-$th$ round.

\par Besides, we also need to constantly estimate and update the current distribution function in every round. For example, if we have evaluation results $\boldsymbol{y}=\left\{y_i| i=1,...,\sum{N(j)}\right\}$, the distribution can be assumed as a non-parameterized function as,
\begin{equation}
	f_{D_{r}}(\boldsymbol{x})=\dfrac{1}{h^{d}\sum_{r}{N(r)}}\sum_{r}\sum_{i=1}^{N(r)}y_{i}\phi(\dfrac{\boldsymbol{x}-\boldsymbol{x}_{i}}{h})
\end{equation}
where $D_r$ is the distribution after $r$-$th$ testing, $h >0$ is a pre-chosen bandwidth, and $\phi(\cdot)$ is a special kernel function.
\par Then we can use new distribution $D_r$ instead of previous $D_{r-1} $to assist the $r$+1$th$ testing.

\subsection{Sampling as Sub-optimization in Inner Loop}
\par The inner loop is embedded in the outer loop, which takes the knowledge learned by the outer loop as the input and outputs the newly sampled sample points in the round. 
\par Similarly, we describe unknown subspace covering problem into a constrained optimization problem. At the $r$-$th$ outer loop, the inner loop receives newly generated samples $\boldsymbol{X}^{(r)}$ as input. Since after the previous r-1 outer loops, the previous samples formulate the known subspace, our job is to maximize the coverage of the reamining unknown subspace using $\boldsymbol{X}^{(r)}$. We define this coverage as $\chi_{ukn}(\boldsymbol{X}^{(r)})$. Thus, the sub-optimization problem is as follows:
\begin{equation}\begin{gathered}
		\label{eq6}
		\operatorname*{max}Volume(\chi_{ukn}(\boldsymbol{X}^{(r)}))   \\
		s.t.\varepsilon_{d}^{i}>0,\forall i,d \\
		s.t.\chi(\boldsymbol{x}_{i})=\{\mid x_{d}^{\prime}-x_{i,d}\mid\leq\varepsilon_{i,d}\mid d=1,2,...D, \boldsymbol{x}_{i} \in \boldsymbol{X}^{(r)}\},\forall i \\
		s.t. \boldsymbol{X}^{(r)} \subseteq \complement_{Space}{\left(\bigcup_{j=1}^{r-1}\chi(\boldsymbol{X}^{(j)})\right)}
\end{gathered}\end{equation}
where $\left(\bigcup_{j=1}^{r-1}\chi(\boldsymbol{X}^{(j)})\right)$ is the known subspace formed by samples generated in previous outer loops and their covering subspaces while $\complement_{Space}{\left(\bigcup_{j=1}^{r-1}\chi(\boldsymbol{X}^{(j)})\right)}$ is the complement of these subspaces, which is unknown subspace in current outer loop.  

\par The objective function is the size of the remaining unknown subspace area. The first condition constrains that the coverage range of a sample must be strictly greater than 0. The second condition limits the coverage of one sample. The third condition reveals that the newly generated samples should be in unknown subspace. 
\begin{figure}[htb]
	\centering
	\includegraphics[width=3.5in,keepaspectratio]{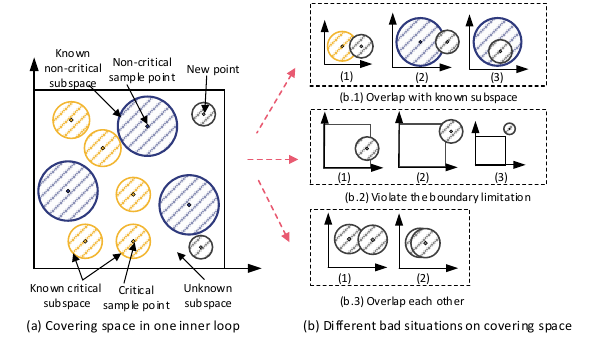}
	\caption{Illustration for different situations on covering space under the two-dimensional projection.}
	\label{Fig4}
\end{figure}
The solution to Eq.\ref{eq6} is helpful to solve Eq.\ref{eq4} because it can accelerate the convergence of the outer loop by covering more unknown subspace. Otherwise, more samples would be generated to cover the space. The reason is that it is difficult to fill a certain number of samples perfectly into the unknown subspace at one time considering the complex and irregular known subspace. Even if the arrangement is regular, the samples will inevitably overlap \cite{buchholz2019improving}.

\par Obviously, we cannot directly solve this optimization problem. We need to iteratively update the positions of the points to find the best sampling scheme, aiming at better covering the unknown subspace. To realize this, we also make the following assumptions and rewrite the problem.
\subsubsection{Assumptions on the coverage of sample}
\par First, we assume that  $\varepsilon_{d}$ for one sample point are unified. Otherwise, the coverage of the sample may be a non convex multidimensional polygon and it will bring difficulties to the rapid solution. Furthermore, the coverage of a sample point can be regarded as a $d$-dimensional sphere \begin{equation}\chi(\boldsymbol{x}_i)=\{\boldsymbol{x}\parallel\boldsymbol{x}-\boldsymbol{x}_{i}\mid^{2}\le(R_{i})^{2},R_{i}>0\},\forall i\end{equation}
where $R_{i}$ is the sphere radius of the $i$-$th$ point.

\par Thus, the known subspace can be seen as a combination of several spheres of different sizes, which may be overlapped, tangent, separated. We assume that the sphere radius depends on the criticality of the sample. When the sample is critical, the sphere radius is relatively small, indicating that its neighborhood needs to be paid more attention for further exploration. On the contrary, when the sample is not challenging, its sphere radius is relatively large. From this assumption, we expect to explore subspace with higher possibility to have critical points due to a limited testing resource. 
\par The process is like packing a set of spheres in box-like space. We will expand it in the next section.

\subsubsection{Modification on the objective function}
\par Refer to \cite{10415067}, we transform the maximization of the coverage volume of the unknown subspace into the maximization of the weighted distance between samples for a quick solution. Define $Dist(\boldsymbol{x}_{i},\cdot)$ as the weighted distance between any two spheres, which can be written as 
\begin{equation}
	Dist(\boldsymbol{x}_{i},\cdot)=\min \left\{\lambda_m Dist(\boldsymbol{x}_i,\boldsymbol{u}_m), \alpha_j Dist(\boldsymbol{x}_i, \boldsymbol{x}_j) \right\} \forall i,j,m
\end{equation}
where $\lambda_m$ and $\alpha_j$ are the two weighted coefficients. 
\par Then the sub-optimization problem in inner loop is rewritten as,
\begin{equation}\begin{gathered}
		\max_{\boldsymbol{x}}{\frac{1}{I}}\sum_{i}Dist(\boldsymbol{x}_{i},\cdot) \\
		s.t.i\leq I \\
		{s.t.}\chi(\boldsymbol{x}_i)=\{\boldsymbol{x}\parallel\boldsymbol{x}-\boldsymbol{x}_{i}\mid^{2}\leq (R^i)^{2}\} \\
		s.t.\chi(\boldsymbol{u}_m)=\{\boldsymbol{u}\mid\mid\boldsymbol{u}-\boldsymbol{u}_m\mid^{2}\leq {(R^m)}^{2}\} \\
		s.t.\boldsymbol{x}_{i}\in\boldsymbol{X}^{(r)},\boldsymbol{u}_{m}\in{\left(\bigcup_{j=1}^{r-1}\chi(\boldsymbol{X}^{(j)})\right)} \\
\end{gathered}\end{equation}
\par The core idea of this modification is to let new spheres move away from known subspace by maximizing the average distance between them. When these new spheres do not overlap with known spheres or with each other, they will certainly cover more unknown subspace. Besides, the calculating of distance between two spheres is much more faster and easier than directly calculating the coverage to unknown subspaces.
\section{The Adaptive Process of Packing Spheres in Inner Loop}
\par We suggest a heuristic strategy that utilizes the concept of $'$anthropomorphism$'$ to visualize the sampling process. Instead of describing sampling as the taking of sample point from a space full of them, we can view it as packing a certain number of spheres in an box-like space. That is, the sampling process can be reversed to place one sample sphere into the space. In this way, the new spheres put into the space may have several undesirable states, as shown in Fig.\ref{Fig4}: the new spheres overlap with each other, the new spheres cover part of the known space, and the new spheres reach over the boundary limit, etc. These states will affect the coverage results.
\par
\begin{figure}[htb]
	\centering
	\includegraphics[width=3.5in,keepaspectratio]{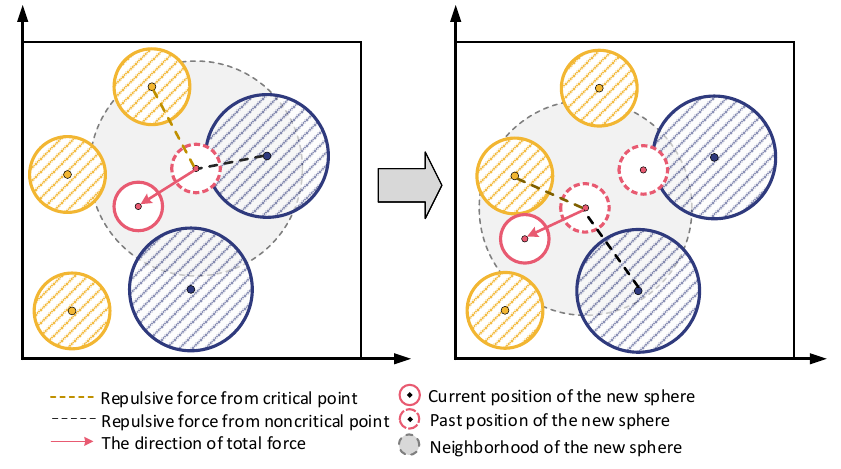}
	\caption{Illustration on the moving process step by step under the two-dimensional projection.}
	\label{Fig5}
\end{figure}

 To alleviate this, we first assume that the known sphere is immovable for all following iterations. Next, we assume the existence of a repulsive force on the sphere.  Then a new sphere, subject to the repulsive force of the neighbors, will move in the opposite direction at each iteration and also actively repel the other newly placed spheres (see in Fig.\ref{Fig5}). A KD-tree is constrcuted to model the relationship between the sample and its neighbors\cite{10006798} and is helpful to search the samples in neighborhood quickly.

\par To better understand this repulsion, we define the factors influencing it are the properties of the spheres and the distance between their centers. The idea is improved by us being inspired by the elegant solution to the packing problem in \cite{LiangYu2012two}\cite{chen2016introduction}. The property of sphere, also called repulsion coefficients, is related  to the criticality of the sample. For example, repulsion coefficients of known critical samples are relatively small, suggesting the possibility of many critical scenarios in the subspace vicinity. Conversely, non-critical samples have a relatively large repulsion coefficient. Also, the repulsion force between two spheres is proportional to the inverse of the square of the distance between their centers. The closer the spheres are, the greater the repulsion force. Spheres will not repel the new sphere if they are outside the preset range of the new one. Based on this, the repulsive force between a known sphere and a new sphere is defined as,

\begin{equation}
		{\overrightarrow{F}}(\boldsymbol{x}_i,\boldsymbol{u}_m)=\frac{\mu Q(\boldsymbol{u}_m)q(\boldsymbol{x}_i)}{\overrightarrow{dist}(\boldsymbol{u}_m,\boldsymbol{x}_i)^2}
\end{equation}
where $Q(\boldsymbol{u}_m)$ is the coefficient of the known sphere, $q(\boldsymbol{x}_i)$ is the coefficient of new sphere, and $\overrightarrow{dist}(\boldsymbol{u}_m,\boldsymbol{x}_i)$ is the distance between the centres of two spheres. Similar, we define the force between new spheres is 
\begin{equation}
	 {\overrightarrow{F}}(\boldsymbol{x}_i,\boldsymbol{x}_j)=\frac{\mu q(\boldsymbol{x}_j)q(\boldsymbol{x}_i)}{\overrightarrow{dist}(\boldsymbol{x}_j,\boldsymbol{x}_i)^2}
\end{equation}.
\par Besides, to prevent the sphere moving outside the boundary, the log barrier function is used to define the binding force subject to the $d$-$th$ dimension,

\begin{equation}
	{\overrightarrow{F'}_d}(\boldsymbol{x}_i)=-\beta\log(\mathrm{Prj}_{{d}}(\pmb{x}_{i}^{(k)})){\overrightarrow{1}}_{d}
\end{equation}
where $\beta$ is the weight used to balance the repulsive force and the binding force from the boundary.

After k iterations in a inner loop, the position of the points $\boldsymbol{x}_i$ is expressed as $\pmb{x}_{i}^{(k)}$. And at the next iteration, we can calculate the combined force as follows,
\begin{equation}
	\begin{aligned}
		&{\overrightarrow{F}}_{total}(\pmb{x}_{i}^{(k+1)})=\sum_{m=1}^{M}\lvert F(\pmb{x}_{i}^{(k)},\pmb{u}^{m})\rvert\frac{\overrightarrow{dist}(\pmb{x}_{i}^{(k)},\pmb{u}_{m})}{\lvert\overrightarrow{dist}(\pmb{x}_{j}^{(k)},\pmb{u}_{m})\rvert}\\
		&+\sum_{j=1}^{I-1}\lvert F(\pmb{x}_{i}^{({k})},\pmb{x}_{j}^{(k)})\rvert\frac{\overrightarrow{dist}(\pmb{x}_{i}^{(k)},\pmb{x}_{j}^{(k)})}{\lvert\overrightarrow{dist}(\pmb{x}_{j}^{(k)},\pmb{x}_{j}^{(k)})\rvert}\\
		&+\sum_{d=1}^{D}-\beta\log(\mathrm{Prj}_{{d}}(\pmb{x}_{i}^{(k)})){\overrightarrow{1}}_{d}\\
	\end{aligned}
\end{equation}

\par Given the huge size of the number of samples, we only count the sphere with the neighboring spheres in the space. The reason is that it has less influence on the change of the direction of the sphere movement when the repulsive force is relatively small. Besides, traversing all the spheres will bring a very high computational cost. The principle of KD-tree is to divide the space into hyper-rectangular regions based on the positions of all the spheres and represent the sphere center coordinates as tree nodes. Then the neighboring samples of each sample can be quickly found by tree search.

\par Note that there is a special case where the updated sphere may still overlap with the known subspace or other spheres when moving in the direction guided by the combined force. To address this, a probability function is used to determine whether to accept the new position at current iteration. If the new solution is worse than the previous iteration's solution, the probability of accepting it decreases exponentially with the number of iterations. Conversely, if the new solution is better than the previous iteration's solution, the probability of updating is $1$.
\begin{equation}
	\begin{gathered}
		\delta_{k+1}=\left\{\begin{array}{c}{1,if\frac{1}{I}\sum_iDist(\boldsymbol{x}_i^{(k+1)},\cdot)>\frac{1}{I}\sum_iDist(\boldsymbol{x}_i^{(k)},\cdot),\forall i}\\{e^{\psi(k)}, otherwise}\end{array}\right.
	\end{gathered}
\end{equation}
where 
\begin{equation}
	\psi(k)=-\frac{\tau(k)\cdot\frac{1}{I}\sum_iDist(\boldsymbol{x}_i^{(k)},\cdot)-\frac{1}{I}\sum_iDist(\boldsymbol{x}_i^{(k+1)},\cdot)}{[\frac{1}{I}\sum_iDist(\boldsymbol{x}_i^{(k)},\cdot)-\frac{1}{I}\sum_iDist(\boldsymbol{x}_i^{(k+1)},\cdot)]}
\end{equation}
, and $\tau(k)$ is an increasing function.
\par So the position of $i$-$th$ point at $k$+1-$th$ iteration in one inner loop is 
\begin{equation}
	\begin{gathered}
		\boldsymbol{x}_i^{(k+1)}=\left\{\begin{array}{c}{\boldsymbol{x}_i^{(k)}-t_{k+1} \frac{\overrightarrow{F}_{total}(\boldsymbol{x}_i^{(k)})}{|\overrightarrow{F}_{total}(\boldsymbol{x}_i^{(k)})|},if \quad p_{k+1} < \delta_{k+1}}\\{\boldsymbol{x}_i^{(k)}, otherwise}\end{array}\right.
	\end{gathered}
\end{equation}
where $p_{k+1}$ is the value generated randomly in $(0,1)$. 

Thus, we provide the algorithm for the life-long learning and testing.
\begin{algorithm}[h]
	\label{alg1}
	\caption{Life-long learning and testing}
	\LinesNumbered 
	
	\KwIn{ $Behavior$ $models$; $r=0$, $N(r)$; $D$; $R_i$;$Score_{\max}$;$CRate_{\theta}$ Testing metrics}
	\KwOut{$Score(\boldsymbol{X})$; $\boldsymbol{X}$; \textit{Space knowledeg}}
	
	\textit{Outer loop}:
	
		\textbf{Construct sampling space}
		
		\While{$CRate<CRate_{\theta}$} 
		{
			\textbf{Generate }set $\boldsymbol{X}^{(r)}$ randomly with $N(r)$ new points 
			
			\textbf{Run} \textit{Inner Loop}
			
			\textbf{Evaluate} critical scenarios and \textbf{Update} $Score(\boldsymbol{X})$
			
			\textbf{Update} space knowledge 
			
			$r += 1$
		}

\end{algorithm}

\begin{algorithm}[h]
	\label{alg2}
	\caption{Adaptive sampling strategy for inner loop}
	\LinesNumbered 
	
	\KwIn{ $\boldsymbol{X}^{(r)}$;Space knowledge; $D$; $R^i$;$K$ for total iteration steps; $\delta$, $n$}
	\KwOut{Updated points $\boldsymbol{X}^{(r)}$}
	\While{$k<K$}{
		\textbf{Calculate} $Dist(\boldsymbol{X}^{(r)},\cdot)$
		
		 $TempDist = \frac{1}{I}{Dist(\boldsymbol{X}^{(r)} ,\cdot)}$
		 
		\textbf{Select} $n$ points randomly
		
		\For{$\boldsymbol{x}^{(k)}$ in selected $n$ points}{
		
			\textbf{Search} neighbors
			
			\textbf{Calculate} $\overrightarrow{F}_{total}(\pmb{x}^{(k)})$ and direction
			
			\textbf{Calculate} the next position for point

	}
				
	\textbf{Calculate} $Dist(\boldsymbol{X}^{(r)},\cdot)$
	
		\If{$\frac{1}{I}Dist(\boldsymbol{X}^{(r)},\cdot)>$ $TempDist$}{
		
			\textbf{Update} the new positions
			
			$TempDist = \frac{1}{I}{Dist(\boldsymbol{X}^{(r)} ,\cdot)}$
			
		}
		\Else{
		\textbf{Update} the new positions with  $\delta_k$ 
		
		}

	$k += 1$
	}

\end{algorithm}
\begin{figure}[htb]
	\centering
	\includegraphics[width=3.5in,keepaspectratio]{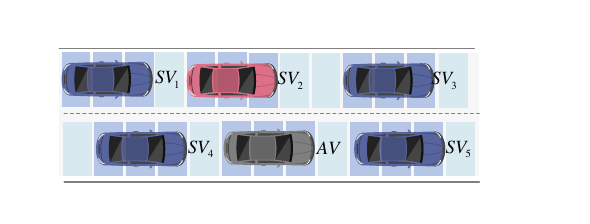}
	\caption{Settings on the static environment.}
	\label{Fig6}
\end{figure}

\section{Simulation Results}
\label{sec5}
\subsection{Simulation Settings}
\label{sec5_A}
\par As shown in the Fig.\ref{Fig6}, the static environment of the generated atom scenario is unidirectional two-lane test road, while the traffic participants are six vehicles in each of the two lanes, including five SVs and one tested AV. In this simulation, only the surrounding vehicles can perform lane changing.
\begin{figure*}[]
	\centering
	\includegraphics[width=5.1in,keepaspectratio]{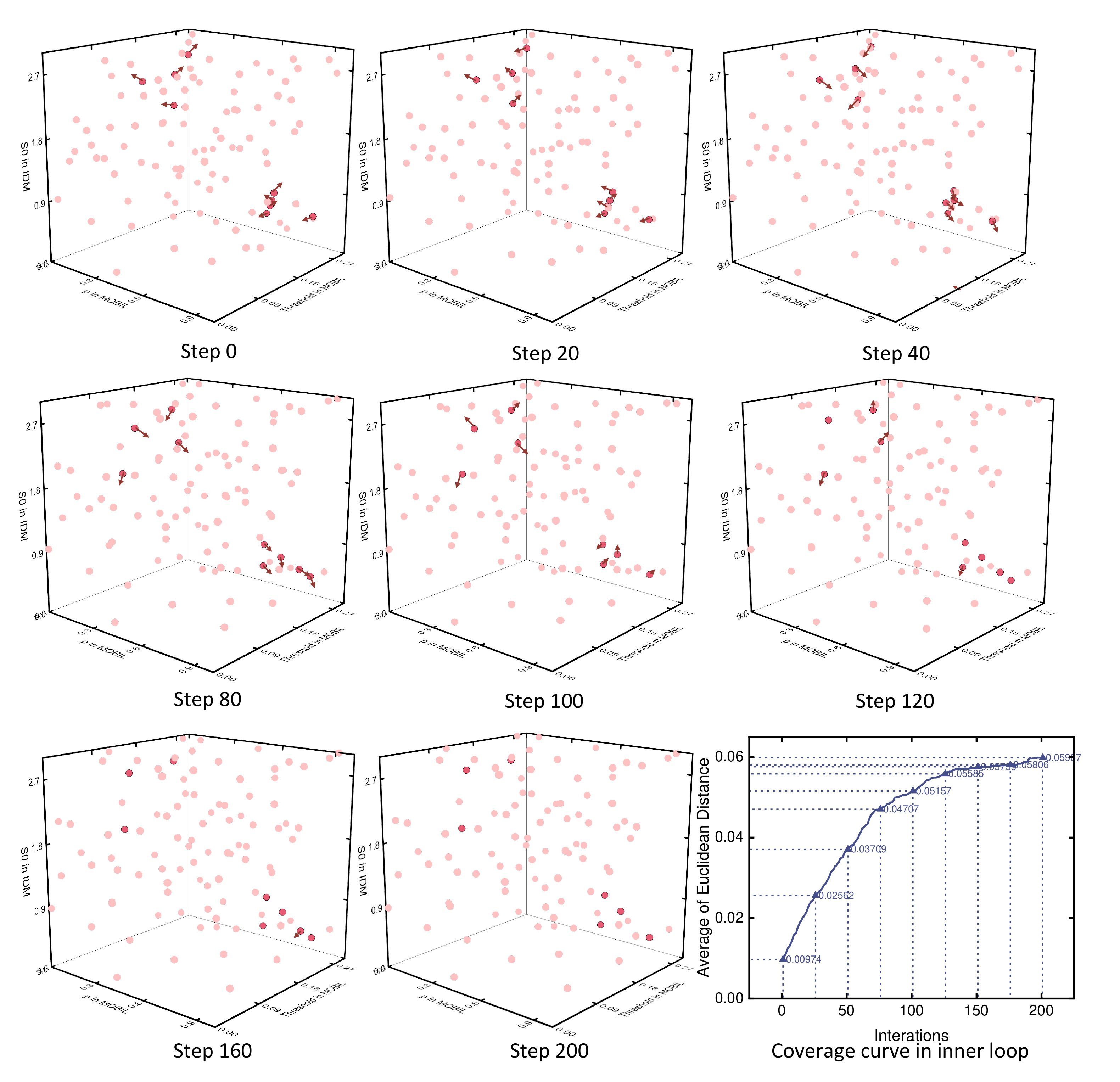}
	\caption{Moving spheres at different iterations in inner loop.}
	\label{Fig7}
\end{figure*}

\par Referring to \cite{10415067}, we adopt the IDM, a model with a describable composition of parameters, to simulate car-following behavior of the SVs. IDM model has five main parameters: $v_0$  is the free stream velocity of the vehicle, $\alpha$  is the maximum acceleration, $s_0$ is the minimum distance in congested traffic, $T$ is a constant safe time gap, and $b$ is the 'comfortable deceleration'. In addition, we use MOBIL as the lane-changing behavior model, which has two main parameters: $p$ and $a$. $p$ stands for 'politeness', indicating the degree of politeness of the surrounding vehicles to change lanes while $\Delta a_{th}$ represents the threshold for the change of acceleration. The following are the representations of these two models, respectively:
\begin{equation}
	\textit{IDM}: a^i=\alpha\left[1-\left(\frac{v^i}{v_0}\right)^4-\left(\frac{s^*\left(v^i, \Delta v^i\right)}{s^i}\right)^2\right].
\end{equation}

\begin{equation}
\textit{MOBIL}:\underbrace{\tilde a_c-a_c}_{\mathrm{SV}}+p\bigl(\underbrace{\tilde a_n-a_n}_{\mathrm{new~follower}}+\underbrace{\tilde a_o-a_o}_{\mathrm{old~follower}}\bigr)>\Delta a_{\mathrm{th}}
\end{equation}
where 
\begin{equation}
	s^*\left(v^i, \Delta v^i\right) = s_0+vT+\frac{v^i\Delta v^i}{2\sqrt{ab}}
\end{equation}

\par It is also necessary to point out that our approach is more concerned with the critical scenarios where SVs play reasonable and rational, i.e., these drivers do not aim to create accidents.

\par In this paper, the AV is required to safely and efficiently finish car-following task while SV cuts in. Since we only test the performance of AV, it's not necessary to model the whole characteristics of it. Instead, a collision avoidance model can be used as an alternative model to mimic the AV to keep a reasonable distance from the leading vehicle, if its parameter $G$ is modified properly\cite{meng2017analysis}\cite{kometani1958stability}. Thus, the following speed of AV at the next moment is calculated as follows
\begin{equation}
	v^{AV}\left(t+\Delta t\right)=\begin{cases}\max\left\{0,v^{AV}(t)-a_{\max}\cdot\Delta t\right\},L(t)<G\\\min\left\{v_{\max}^{AV},v^{AV}(t)+a_{\max}\cdot\Delta t\right\},L(t)\geq G\end{cases}
\end{equation}
where desired $G$ is set to $[6,12]m$ in this paper.

\begin{figure*}[htb]
	\centering
	\includegraphics[width=4.5in]{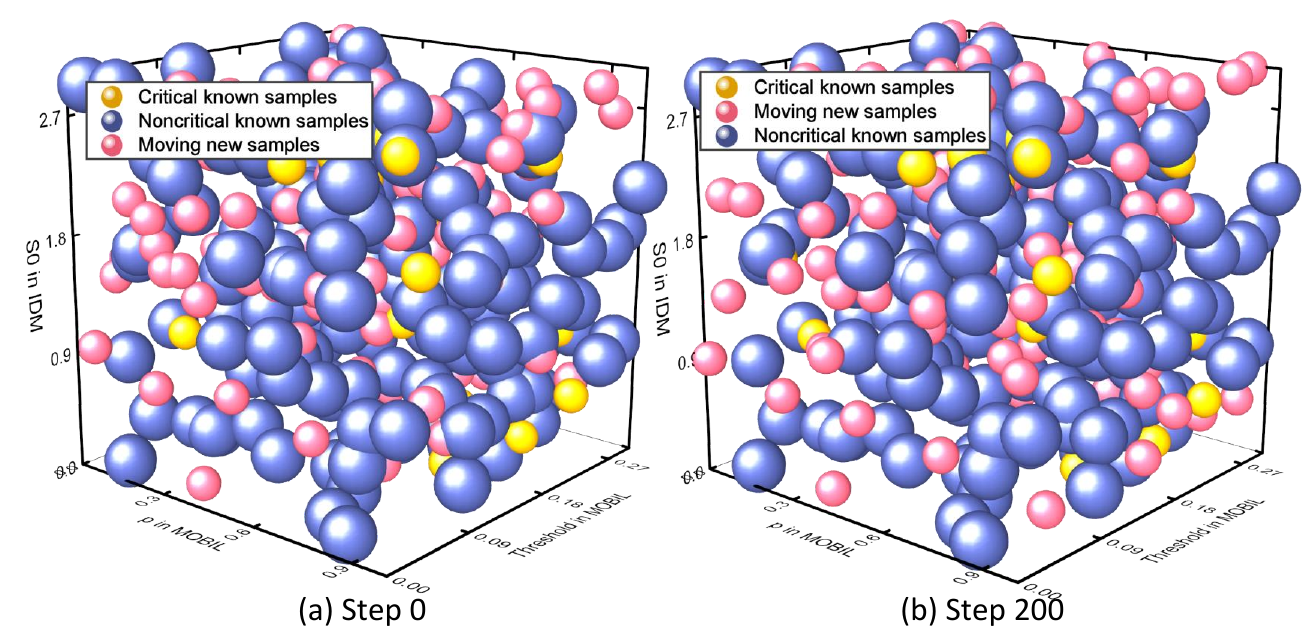}
	\caption{Illustrations on the start and end of the spheres packing in inner loop. After 200 iterations in inner loop, it can be vividly seen that the new samples (red spheres) are better explored in unknown subspace, from aggregated to dispersed.}
	\label{Fig8}
\end{figure*}

\begin{table*}[!htb]
	
	\footnotesize
	\renewcommand{\arraystretch}{1.3}
	\centering
	\caption{Average of Euclidean Distance}
	\centering
	\setlength{\tabcolsep}{4.5mm}{
		\begin{tabular}{cccccc}
			\hline\hline
			New Samples & 200& 600 & 1000 & 1400 & 2000\\ \hline
			Greedy & $-7.81*10^{-4}$& $-5.66*10^{-4}$ & $-6.93*10^{-4}$ & $-7.02*10^{-4}$ & $-7.06*10^{-4}$\\
			Monte Carlo & $-1.49*10^{-2}$& $-1.59*10^{-2}$ & $-1.61*10^{-2}$ & $-1.64*10^{-2}$ & $-1.70*10^{-2}$\\
			Ours & $0.0134$& $0.0115$ & $0.0826$ & $0.0059$ & $0.0034$\\
			\hline
			\label{tbl2}
	\end{tabular}}
\end{table*}
\subsection{The Construction of Sampling Space}

\par We consider that a scenario begins when the parameters of models update and ends when $SV_2$ completes a lane change behavior. It is because  $SV_2$ is most likely to put AV into accidents. Besides, the parameters remain constant in one scenario otherwise the size of the sampling space increases rapidly as the simulation time grows. Based on the literature \cite{alhariqi2022calibration}, the upper and lower bounds of the behavioral model parameters of one SV are set as shown in the TABLE.\ref{Table_1}.
\par All the simulations are conducted in the CAVsim platform. CAVsim supports a wide range of vehicle behavior modeling and can output the testing results such as TTC in real time. 
\begin{table}[!htb]
	\footnotesize
	\renewcommand{\arraystretch}{1.3}
	\centering
	\caption{Bounds of Parameters.}
	\centering
	\setlength{\tabcolsep}{4.5mm}{
		\begin{tabular}{ccc}
			\hline\hline
			Symbol & Lower Bound& Upper Bound\\ \hline
			$v_0$ & $25 m/s$ & $30 m/s$\\
			$\alpha$ &  $1 m/s^{-2}$ &  $5 m/s^{-2}$\\
			$T$ & $0.05s$ & $2s$\\
			$b$ & $0.1 m/s^{-2}$ & $4 m/s^{-2}$\\
			$s_0$ &  $0.1m$  & $3$\\
			$p$ &  $0$ & $1$\\
			$\Delta a_{th}$ &  $0 m/s^{-2}$ & $0.3 m/s^{-2}$\\
			\hline
	\end{tabular}}
	\label{Table_1}
\end{table}

\subsection{Results on Adaptive Sampling in Inner Loop}
\par This section demonstrates the process of iterating and moving the sample spheres in the 3-dimensional parameter space. It also examines the impact of two important parameters, the number of samples and the radius of the spheres. Finally, the algorithm with our strategy is compared to other classical algorithms to highlight its superiority.
\begin{figure}[htb]
	\centering
	\includegraphics[width=2.5in]{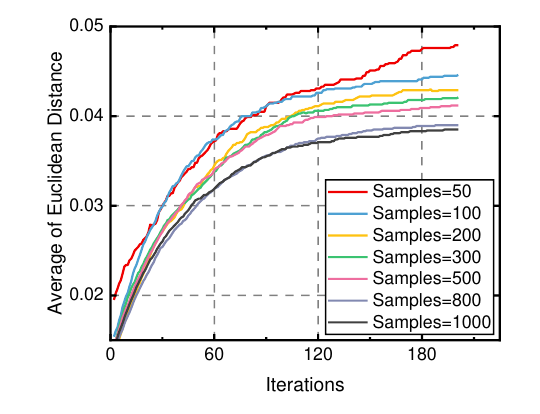}
	\caption{Comparison results on simulating different new samples in one inner loop.}
	\label{Fig9}
\end{figure}
\begin{figure}[htb]
	\centering
	\includegraphics[width=2.5in]{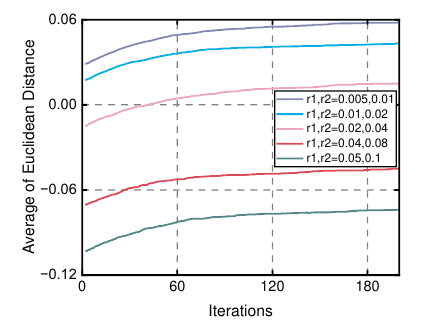}
	\caption{Comparison results on simulating spheres with different sizes in one inner loop.}
	\label{Fig10}
\end{figure}
\par Fig.\ref{Fig7} displays the movement of 100 new spheres among 200 known spheres throughout the inner loop. Three subspaces are selected and we deepen the color of the spheres in them. The dark red sphere continues to move during the iteration process and stabilizes after 160 iterations, indicating that the optimal position has been found. In Fig.\ref{Fig8}, we added the known subspace and compared the inner loop's initial and final states.
\par Based on Fig.\ref{Fig7} and Fig.\ref{Fig8}, it can be analyzed that in order to cover more unknown space, the new spheres keep moving away from the known subspace due to the repulsive force. To achieve a balance between high coverage and speed in identifying critical samples, the strategy let spheres explore the unknown subspaces near the critical samples first. We can see that new spheres are moved away from non-critical known subspaces as much as possible and moved away from critical subspaces with a relatively small repulsive force.

\par To analyze the impact of the number of samples and the radius of the spheres. One thousand samples are randomly generated to form a known subspace, with $60$ of them designated as known critical samples. Fig.\ref{Fig9} illustrates the coverage comparison of different numbers of new samples in one inner loop. The results indicate that the higher the number of samples we generate, the smaller the average Euclidean distance.
Under our strategy, increasing the number of samples can better cover the unknown space. However, if the number of samples is too high, the spheres will overlap heavily, resulting in duplicated coverage of the space. Fig.\ref{Fig10} illustrates the change in the average Euclidean distance under different sphere radius settings. When the size of the sphere is larger, meaning that our tolerance to the dissimilarity of neighboring scenes in space is higher, the coverage of the same number of sampling points improves.

\par To verify the coverage effect of the algorithms based on our strategy in the inner loop, we used the Monte Carlo-based randomized generation algorithm and the greedy-strategy-based algorithm as benchmarks. Monte Carlo algorithm is the most commonly used method for scenario generation. It generates sample points directly in the sampling space without relying on any prior information about the subspace. The greedy-strategy-based algorithm is a commonly used for optimization problems. The algorithm repeats the random generation of a certain samples step by step, calculates and compares the coverage metric, and selects the optimal solution of the step as the result. The algorithm stops when the required number of sample points is achieved.

\par In this experiment, the sphere radius for the critical sphere is set to 0.04, while the sphere radius for the non-critical sphere is set to 0.06. A total of 2000 samples were randomly generated to form the known subspace, with 150 of them designated as known critical samples. TABLE.\ref{tbl2} presents the comparison results of different algorithms for generating different number of new samples. Our algorithm provides better coverage of the remaining unknown space, while random sampling yields the worst coverage results. Though the greedy strategy can ensure obtaining a locally superior solution in each iteration, it may not lead to a globally superior solution.

\begin{table*}[]
	\caption{THE NOMENCLATURE LIST}
	\begin{tabular}{@{}cl@{}}
		\hline
		\toprule
		
		SYMBOLS IN THE SCHEME                & \multicolumn{1}{c}{MEANINGS}                                                                \\ \midrule
		$\boldsymbol{X}$                 & The set of scenarios                                                                        \\
		$\boldsymbol{x}_{i}$             & The $i$-$th$ scenario                                                                           \\
		$x_{i,d}$                        & The $d$-th variable of $i$-$th$ scenario                                                        \\
		$\varepsilon_{i,d}$              & A constant that constraints the coverage range of $i$-$th$ sample on the $d$-$th$ dimension \\
		$\chi(\boldsymbol{x}_i)$         & The covering subspace of $\boldsymbol{x}_i$                                                 \\
		$TTC_\theta$                     & Threshold for $TTC$                                                                          \\
		$a_\theta$                       & Threshold for $a$                                                                           \\
		$I\left(\boldsymbol{x}\right)$   & Indicator function to show the criticality of scenario $\boldsymbol{x}$                     \\
		$Score(\boldsymbol{X})$          & The evaluation result of AV passing scenarios $\boldsymbol{X}$                              \\
		$CRate(\boldsymbol{X})$          & The proportion of space covered by the obtained samples                                     \\
		$CRate_\theta$ & Thershold for covering area                                                                 \\
		r                                & Rouond number for outer loop                                                                \\
		$N(r)$                           & Total number of samples in $r$-$th$ round                                                   \\
		$y_i$                            & Evaluation of AV in $\boldsymbol{x}_{i}$                                                    \\
		$D_r$                            & Distribution after $r$-$th$ testing                                                         \\
		$h$                              & Pre-chosen bandwidth                                                                        \\
		$\phi(\cdot)$                    & Special kernel function                                                                     \\
		$R_i$                            & Sphere radius of the $i$-$th$ point                                                         \\
		$Dist(\boldsymbol{x}_{i},\cdot)$ & Weighted distance between any two spheres                                                   \\
		$\lambda_m$                      & 1st Weighted coefficient for $Dist$                                                         \\
		$\alpha_j$                       & 2nd Weighted coefficient for $Dist$                                                         \\
		$\boldsymbol{u}_m$               & $m$-$th$ known scenario                                                                     \\
		$\overrightarrow{F}$             & The force between two spheres                                                               \\
		$\mu$                            & Coefficient for the force                                                                   \\
		$Q(\boldsymbol{u}_m)$            & Repulsion coefficient of the known sphere                                                   \\
		$q(\boldsymbol{x}_i)$            & Repulsion coefficient of the new sphere                                                     \\
		$\overrightarrow{dist}$          & Distance between the centres of two spheres                                                 \\
		$\beta$                          & Weight used to balance the repulsive force and the binding force from the boundary          \\
		$k$                                & Step number in inner loop                                                                   \\
		$\overrightarrow{F}_{total}$     & The combined force                                                                          \\
		$M$                              & Number of neighbors from known samples                                                      \\
		$I$                               & Number of all new samples in one inner loop                                                 \\
		$\delta_{k}$                     & Probability of updating samples in inner loop                                               \\
		$\alpha(k)$                      & An increasing function for controlling the update probability                               \\
		$p_{k}$                          & Randomly generated value                                                                    \\
		$n$                              & Number of samples randomly chosen to move in one outer loop  in ALGORITHM.\ref{alg2}                              \\ 
		\hline
		\toprule
		SYMBOLS IN SIMULATIONS           & \multicolumn{1}{c}{MEANINGS}                                                                \\ \midrule
		$v_0$                            & Free stream velocity of the vehicle                                                         \\
		$\alpha$                         & The maximum acceleration                                                                    \\
		$T$                              & A constant safe time gap                                                                    \\
		$b$                              & Comfortable deceleration                                                                    \\
		$s_0$                            & The minimum distance in congested traffic                                                   \\
		$p$                              & Parameter that represents'politeness'                                                       \\
		$\Delta a_{th}$                  & Threshold for the change of acceleration                                                    \\
		$G$                              & Desired following distance for AV                                                           \\ \bottomrule
	\end{tabular}
\end{table*}

\subsection{Results in Outer Loop}
\begin{figure}[htb]
	\centering
	\includegraphics[width=2.7in,keepaspectratio]{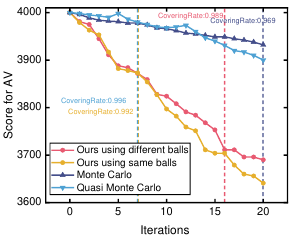}
	\caption{Results on scoring AV. We generate the same number of samples for the benchmarks as the total number of samples in our scheme after each iteration, e.g., we have generated a total of 1600 samples in 8 iterations and we also generate 1600 samples using SMC and QMC respectively.}
	\label{Fig11}
\end{figure}
\begin{figure}[htb]
	\centering
	\includegraphics[width=2.8in,keepaspectratio]{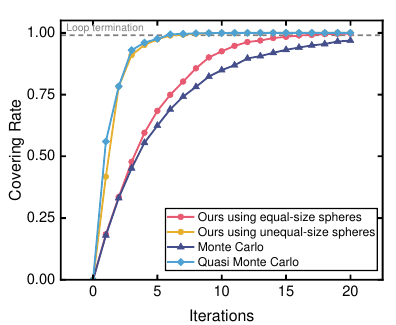}
	\caption{Results on covering rate in outer loop.}
	\label{Fig12}
\end{figure}
\par During the simulation of the outer loop, each loop generates 200 new samples, evaluates the criticality of each scenario, and assigns AV testing score. In this experiment, we set the max Score $S$ is 4000 which corresponds to the total number of generated scenarios. To demonstrate the validity of this scheme, we compare it with standard Monte Carlo (SMC) \cite{broadhurst2005monte}\cite{eidehall2006threat}\cite{eidehall2008statistical}and Random Quasi Monte Carlo (QMC)\cite{feil2009comparison}]\cite{zhou2013mixture}. 
\par But unlike our scheme, none of these methods are iterative. Therefore when presenting the results, we will utilize these methods to generate different numbers of samples independently and ensure that the number of these samples is consistent with the total number of samples generated by our scheme over progressive rounds.
\par Additionally, we conduct another comparison experiment where the space is covered with spheres of the same volume, i.e., we do not distinguish between critical and non-critical samples in terms of sphere radius.

\par The results on AV score are presented in Fig.\ref{Fig11}. It is evident that until the coverage ratio stabilizes, the SMC and QMC methods sample only a few critical samples. This outcome is consistent with previous research \cite{10415067}. In contrast, our method encounters a growing number of critical samples, leading to a significant decrease in AV score. The Fig.\ref{Fig11} displays vertical line segments that indicate that the coverage metric approaches 1, meaning the space is almost completely covered. The method that uses different volumes of sample spheres converges slower but find more critical samples compared to the method that equal-sized spheres.
\par Furthermore, Fig.\ref{Fig12} illustrates that SMC has low coverage and also struggles to quickly sample critical scenarios. On the other hand, QMC has exceptional coverage but performs poorly in quickly finding critical scenarios. Combining Fig.\ref{Fig11} and Fig.\ref{Fig12}, our approach can obtain more critical scenarios and thus test the AVs more accurately, although it sacrifices some coverage. Besides, it is important to note that we only use one kind of coverage metric as a termination condition for the outer loop. We expect further exploration can be conducted to use a better coverage metric.

\section{Conclusion}
\par This paper proposes a novel paradigm to test and improve the intelligence of AV. Most previous studies aim to minimize the expectation of the failing frequency of an AV in various scenarios. 
\par As a result, an AV may be taken as "intelligent enough" even if it cannot successfully pass a certain kind of scenarios which are rare to occur in practice. Differently, we aim to fathomed all the possible scenarios that could be sampled to ensure an AV works well in all such scenarios. Because the sampled scenario space is continuous, we cannot exhaustically enumerate all the scenarios. Because the capability of an AV may vary significantly in different scenarios, we cannot evaluate the intelligence of an AV by just sampling a limited number of the scenarios as representatives and test the AV just in these few scenarios. Therefore, we proposed the life-long learning and test scheme to continously test and (possibly) upgrate the intelligence of AVs. This scheme is indeed an implementation of our idea of building general artificial intelligent systems which is called parallel learning \cite{7974888}\cite{10059315}\cite{li2019parallel}. The challenging scenario sampling strategy proposed in this paper prescibes the intelligent system to further improve itself; while this improvement also leads to new sampling procedures and finally forms a closed-loop spiral ascent of intelligent systems. Constrained by the length limit, there are several issues not discussed in details in this paper. In future, we will further investigate other metric of sampling space coverage.

\bibliographystyle{IEEEtran}
\bibliography{IEEEabrv,IEEEexample}

\begin{IEEEbiography}[{\includegraphics[width=1in,height=1.5in,clip,keepaspectratio]{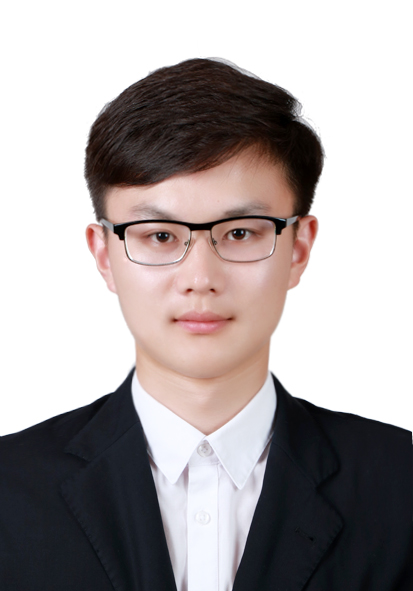}}]{Jingwei Ge}
 is currently pursuing the Ph.D. degree with the Department of Automation, Tsinghua University, China. His current research interests focus on intelligent transportation systems, intelligence testing, and autonomous vehicles testing.
\end{IEEEbiography}
\begin{IEEEbiography}[{\includegraphics[width=1in,height=1.5in,clip,keepaspectratio]{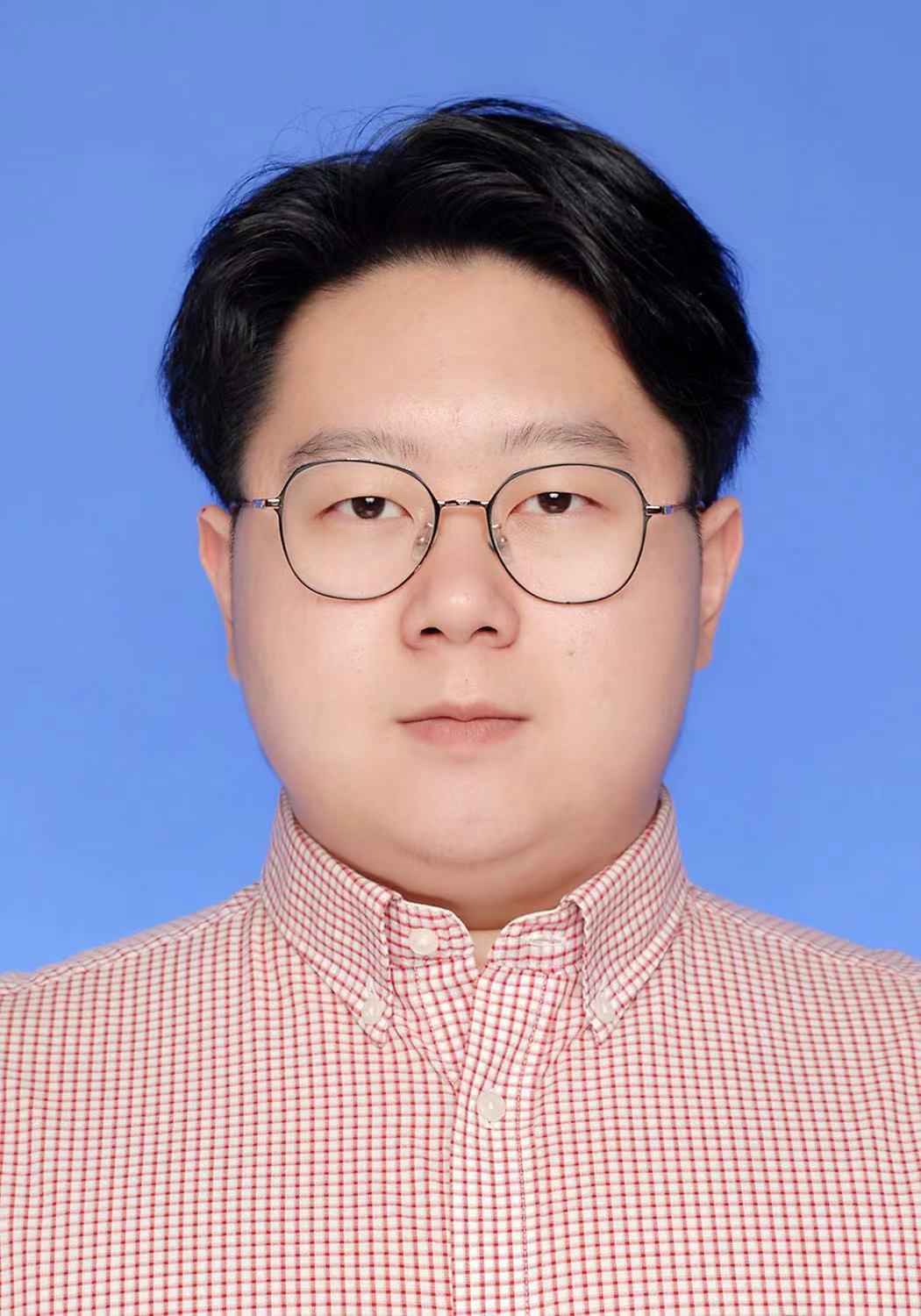}}]{Pengbo Wang}
	is currently pursuing the Ph.D. degree with the Department of Automation, Tsinghua University, China. His current research interests focus on intelligent transportation systems, intelligent vehicles and deep reinforcement learning.
	
\end{IEEEbiography}
\begin{IEEEbiography}[{\includegraphics[width=1in,height=1.5in,clip,keepaspectratio]{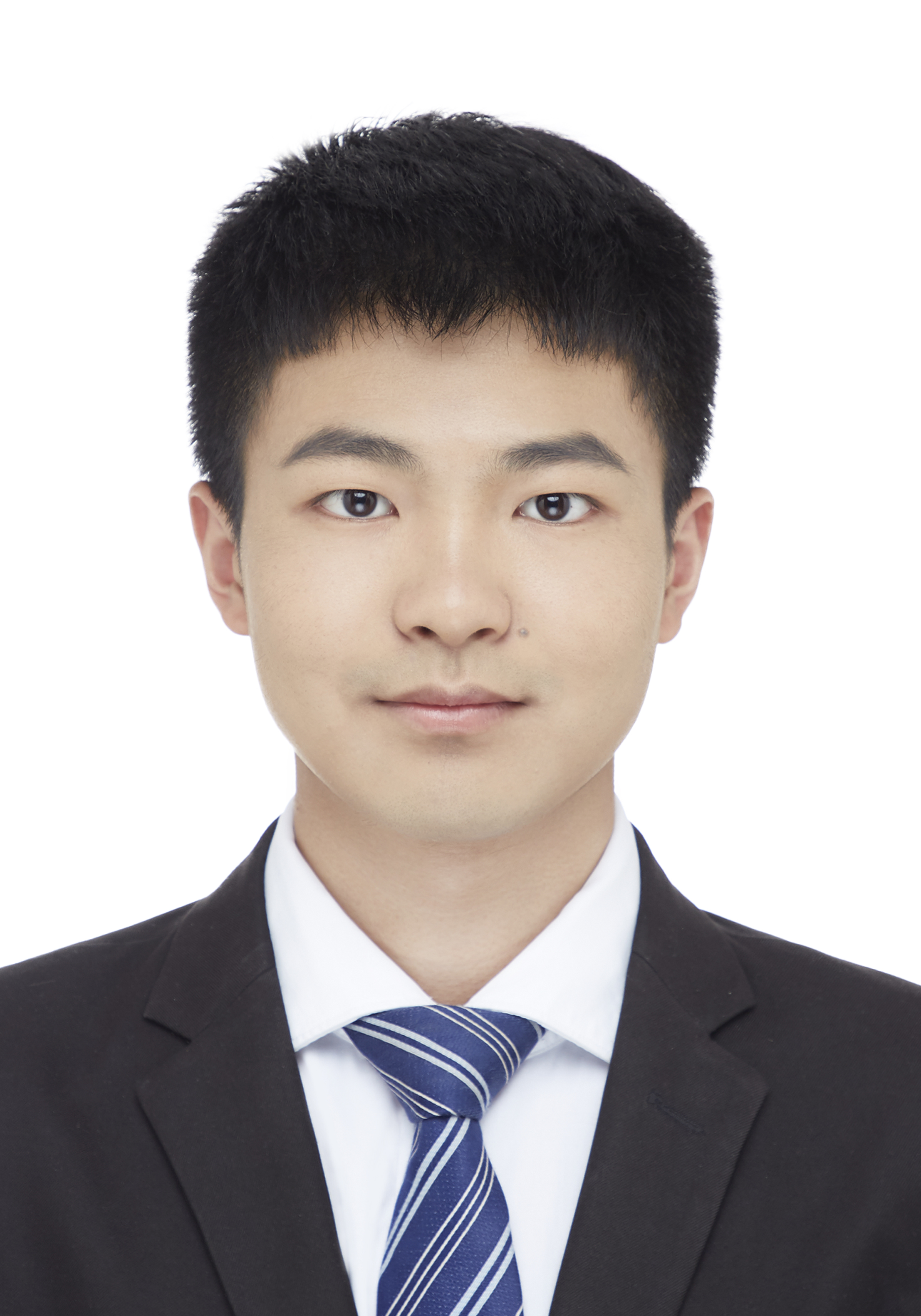}}]{Cheng Chang}
	received the B.S. degree from Tsinghua University, Beijing, China, in 2021, where he is currently pursuing the Ph.D. degree with the Department of Automation. His current research interests include intelligent transportation systems, intelligent vehicles and machine learning.
	
\end{IEEEbiography}

\begin{IEEEbiography}[{\includegraphics[width=1in,height=1.5in,clip,keepaspectratio]{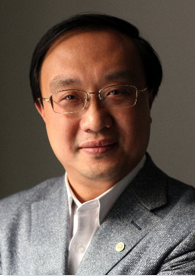}}]{Yi Zhang}
	received the B.S. and	M.S. degrees from Tsinghua University, China, in 1986 and 1988, respectively, and the Ph.D. degree from the University of Strathclyde, U.K., in 1995. He is currently a Professor in control science and engineering at Tsinghua University, with his current research interests focusing on intelligent	transportation systems. His active research areas include intelligent vehicle-infrastructure cooperative systems, analysis of urban transportation systems,	urban road network management, traffic data fusion and dissemination, and urban traffic control and management. His research fields also cover the advanced control theory and applications, advanced detection and measurement, and systems engineering.
	
\end{IEEEbiography}
\begin{IEEEbiography}[{\includegraphics[width=1in,height=1.5in,clip,keepaspectratio]{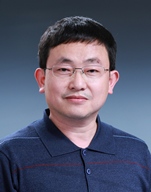}}]{Danya Yao}
	received the B.S., M.S., and Ph.D. degrees from Tsinghua University, Beijing, China, in 1988, 1990, and 1994, respectively. He is currently a Full Professor with the Department of Automation, Tsinghua University. His research interests include intelligent detection technology, system engineering, mixed traffic flow theory, and intelligent transportation systems. 
	
\end{IEEEbiography}

\begin{IEEEbiography}[{\includegraphics[width=1in,height=1.5in,clip,keepaspectratio]{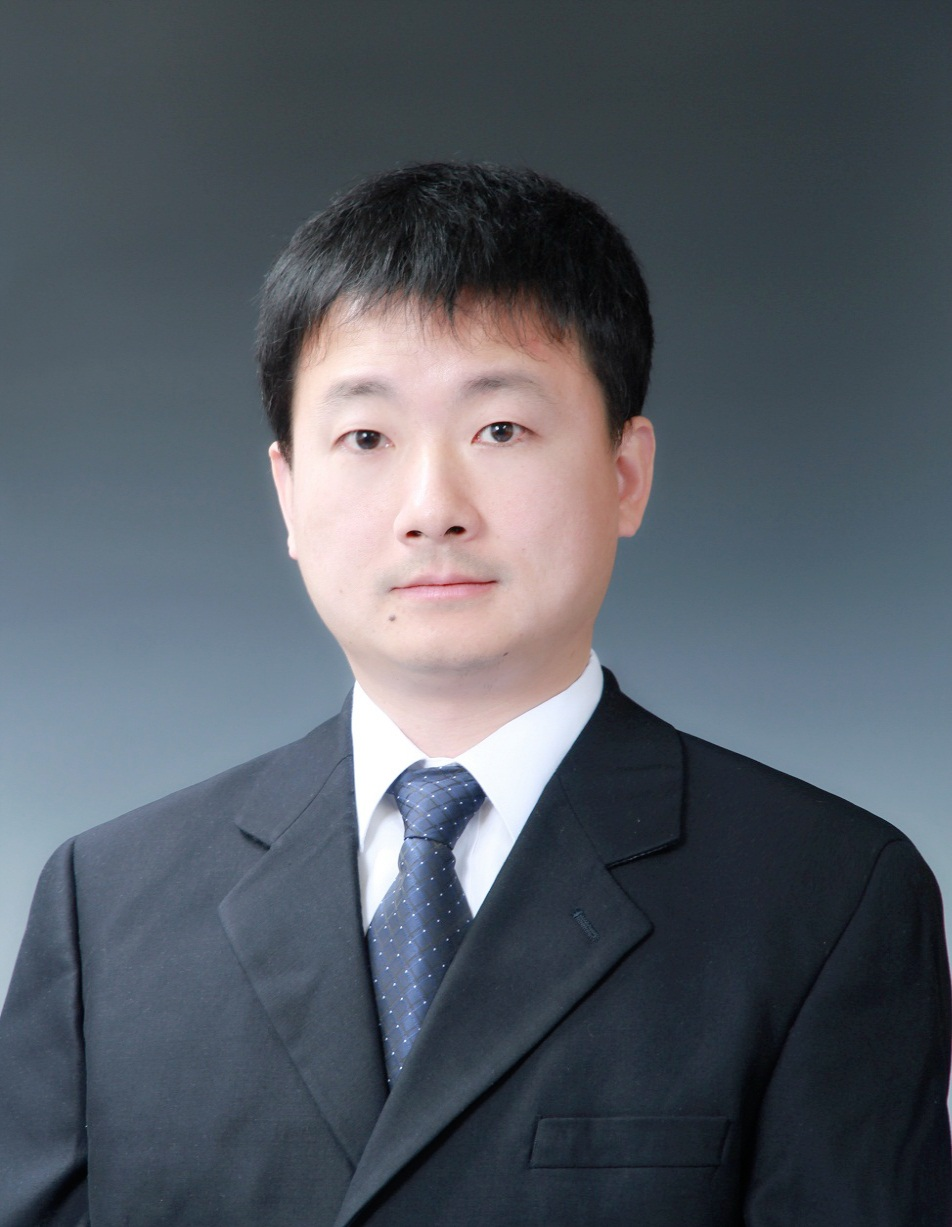}}]{Li Li}
(Fellow, IEEE) is currently a Professor with the Department of Automation, Tsinghua University, Beijing, China, working in the fields of artificial intelligence, intelligent control and sensing, intelligent transportation systems, and intelligent vehicles. He has published over 170 SCI-indexed international journal articles and over 70 international conference papers as a first/corresponding author. He is a member of the Editorial Advisory Board for the Transportation Research Part C: Emerging Technologies, and a member of Acta Automatica Sinica. He also serves as an Associate Editor for IEEE Transactions on Intelligent Transportation Systems and IEEE Transactions on Intelligent Vehicles.
\end{IEEEbiography}

\end{document}